\documentclass[lettersize,journal]{IEEEtran}
\usepackage{amsmath,amsfonts}
\usepackage{algorithmic}
\usepackage{algorithm}
\usepackage{array}
\usepackage[caption=false,font=normalsize,labelfont=sf,textfont=sf]{subfig}
\usepackage{textcomp}
\usepackage{stfloats}
\usepackage{url}
\usepackage{verbatim}
\usepackage{graphicx}
\usepackage{cite}
\usepackage{multirow}
\usepackage[colorlinks=true, citecolor=blue]{hyperref}
\usepackage{xcolor}
\usepackage{booktabs}
\usepackage[numbers, sort]{natbib}
\let\oldcite\cite
\renewcommand{\cite}[1]{\textcolor{blue}{~\oldcite{#1}}}
\usepackage{orcidlink}

\begin{document}

\title{OpenVidVRD: Open-Vocabulary Video Visual Relation Detection via Prompt-Driven Semantic Space Alignment}

\author{Qi Liu~\orcidlink{0000-0001-5378-6404}, ~\IEEEmembership{Senior Member,~IEEE,} Weiying Xue, Yuxiao Wang~\orcidlink{0000-0002-4162-4587}, Zhenao Wei \\ 
\thanks{The authors are with the South China University of Technology, China 511400 (e-mail: drliuqi@scut.edu.cn; 202320163283@mail.scut.edu.cn; ftwangyuxiao@mail.scut.edu.cn; wza@scut.edu.cn).}
}
\markboth{Journal of \LaTeX\ Class Files,~Vol.~14, No.~8, August~2021}%
{Shell \MakeLowercase{\textit{et al.}}: A Sample Article Using IEEEtran.cls for IEEE Journals}

\IEEEpubid{0000--0000/00\$00.00~\copyright~2021 IEEE}

\maketitle

\begin{abstract}
The video visual relation detection (VidVRD) task is to identify objects and their relationships in videos, which is challenging due to the dynamic content, high annotation costs, and long-tailed distribution of relations. Visual language models (VLMs) help
explore open-vocabulary visual relation detection tasks, yet often overlook the connections between various visual regions and their relations. Moreover, using VLMs to directly identify visual relations in videos poses significant challenges because of the large disparity between images and videos. Therefore, we propose a novel open-vocabulary VidVRD framework, termed OpenVidVRD, which transfers VLMs' rich knowledge and powerful capabilities to improve VidVRD tasks through prompt learning. Specificall  y, We use VLM to extract text representations from automatically generated region captions based on the video's regions. Next, we develop a spatiotemporal refiner module to derive object-level relationship representations in the video by integrating cross-modal spatiotemporal complementary information. Furthermore, a prompt-driven strategy to align semantic spaces is employed to harness the semantic understanding of VLMs, enhancing the overall generalization ability of OpenVidVRD. Extensive experiments conducted on the VidVRD and VidOR public datasets show that the proposed model outperforms existing methods.
\end{abstract}


\begin{figure}[]    
  \centering            
  \subfloat[\footnotesize  Conventional VidVRD \textit{\textbf{vs.}} OpenVidVRD.]  
  {  
      \label{fig:subfig1}\includegraphics[width=0.9\linewidth]{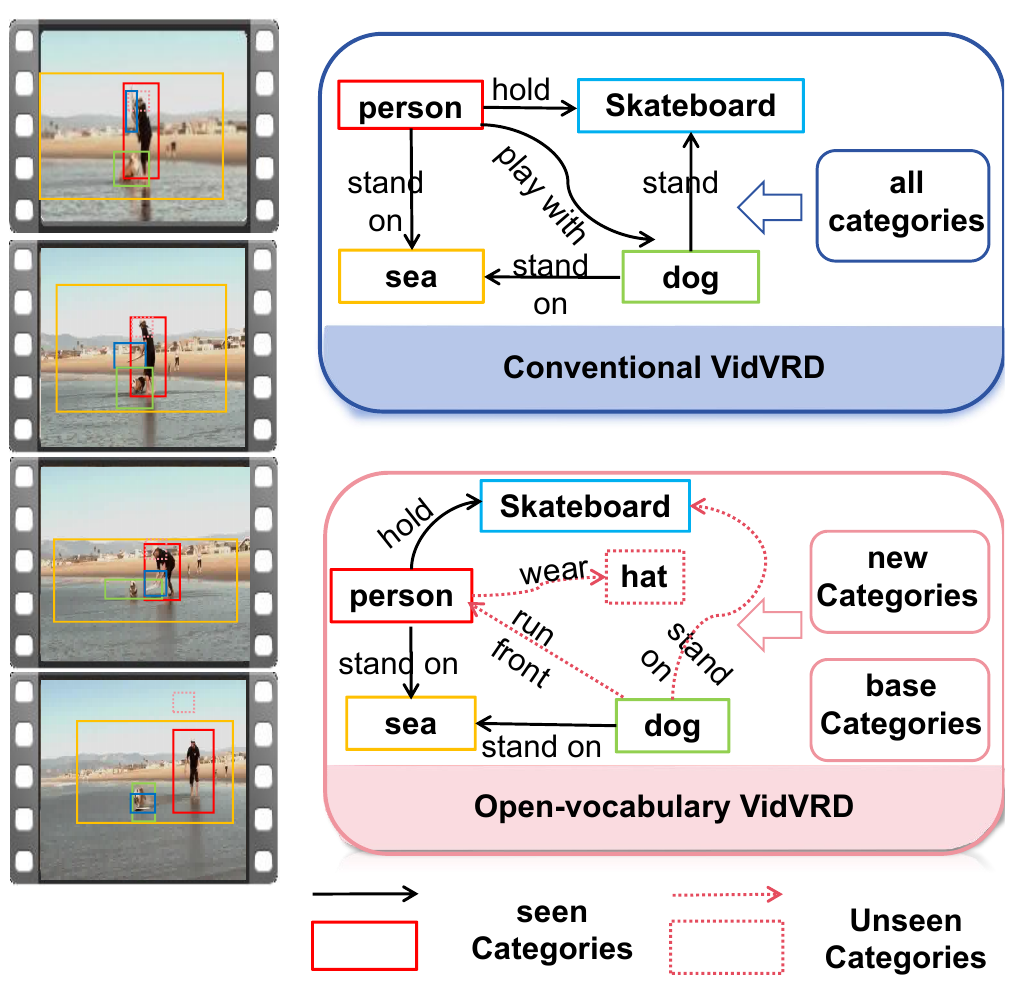}
  }\\
  \subfloat[\footnotesize  The long-tailed distribution of the benchmark dataset VidOR.]
  {
      \label{fig:subfig2}\includegraphics[width=\linewidth]{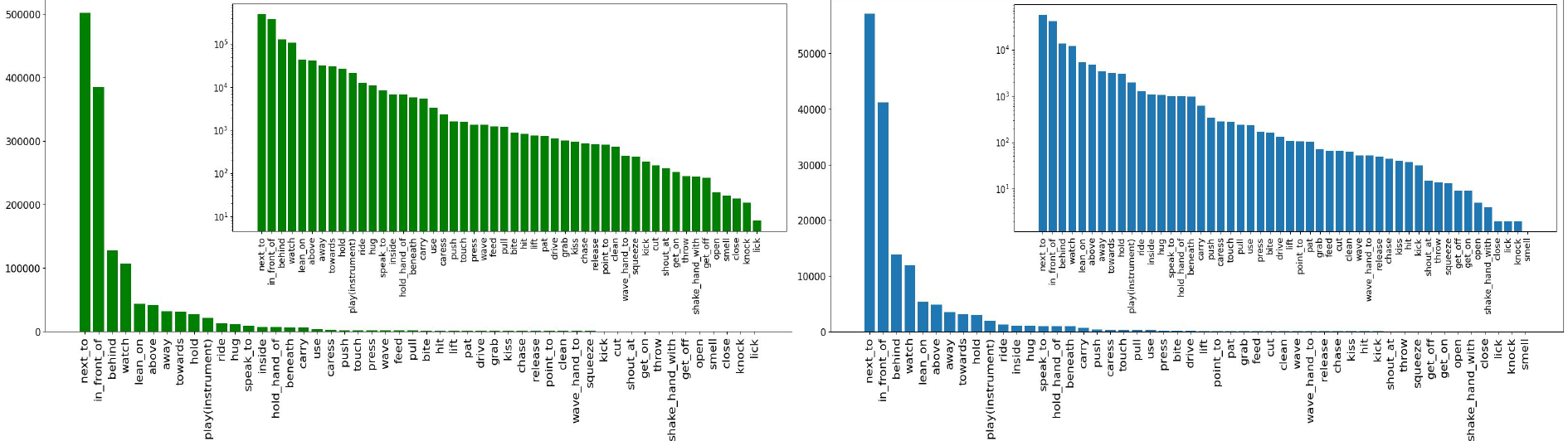}  
  }
  \caption{(a) An illustration of the conventional VidVRD \textit{\textbf{vs.}} Open-vocabulary VidVRD. Conventional VidVRD cannot predict any novel categories, including both predicates and objects. (b) shows the long-tailed distribution of the predicate classes in the VidOR dataset.  }
  \label{intro}          
\end{figure}

\section{Introduction}
\IEEEPARstart{T}{he} main goal of video visual relation detection  (VidVRD) is to identify the intricate visual relations between object tracklets in videos, represented by sets of $\langle$``subject", ``predicate", ``object"$\rangle$ triplets\cite{shang2017video}. It offers a sophisticated understanding and depiction of video data, with various practical uses in tasks, such as scene graph generation\cite{cong2023reltr}, visual question answering\cite{schwenk2022okvqa}, and video understanding\cite{li2023videochat}.

Compared to visual relation detection (ImgVRD) in static images\cite{li2022sgtr,dong2022stacked,lyu2022fine}, VidVRD presents a more challenging task due to the dynamic relations between objects in temporal scales. As shown in Figure~\ref{fig:subfig1}, the verb ``play" can be used in conjunction with other verbs, and also with dynamic relations that evolve over time (for example, ``a person plays with a dog" can be expressed as ``a dog runs towards a person"). The high recall values observed in present methods indicate a possible tendency towards overfitting on head predicate categories (e.g. \textit{in front of/next to}), while not fully assessing the impact in rare classes (e.g. \textit{smell/knock}), particularly those predicates that can provide richer descriptions of potential actions and activities in videos. Figure~\ref{fig:subfig2} illustrates the long-tail distribution of predicate classes in the VidOR dataset. To address these challenges and bridge the gap between fixed-category limitations and real-world relational diversity, recent research has shifted toward Open-Vocabulary Video Visual Relationship Detection (Ov-VidVRD)\cite{gao2023compositional}, where the model was trained on a set of base relations and generalized to infer the novel relation predicates. Recently, the transfer of multimodal knowledge from large-scale pre-training of foundation models for vision-language (V\&L)\cite{li2022blip,radford2021learning,liu2024multi} to various downstream tasks has shown effectiveness and yielded significant 
\IEEEpubidadjcol 
success\cite{gu2021open,liu2024open,zhou2022learning}. Nevertheless, most of them encounter two challenges. The first challenge is generalization on unseen categories. The second challenge stems from addressing the domain gap in relational representations between static image and the spatiotemporal dynamics of video object interactions. Most image-text pretraining models neglect the alignment between visual relationship regions and relational predicate concepts. As indicated by prior studies\cite{yuksekgonul2022and,herzig2023incorporating}, vision-language models (VLMs) like CLIP still encounter challenges in compositional scene understanding, particularly in modeling inter-object relationships. This results in incomplete and unreliable alignment between representations of subject-object pairs and relational predicates, thereby leading to ambiguous relationship predictions.

To that end, a novel open-vocabulary VidVRD framework by integrating prompt-driven semantic space alignment network is proposed, termed OpenVidVRD. Firstly, the VQA model, i.e., Blip2\cite{li2022blip}, is utilized for captioning to align the visual concepts with the corresponding region captions, which can provide more localized semantic relation information. And we introduce a spatiotemporal refiner guided by a visual-text aggregation, enhanced by spatio-temporal visual prompting to adapt CLIP\cite{radford2021learning} for capturing spatial and temporal relationships between objects. Specifically, sequential Transformer blocks\cite{vaswani2017attention} are introduced to model the context among objects, bridging the domain gap in video relationship detection. By focusing on the semantic information of visual regions, our model is further augmented with vision-guided language prompting, which integrates learnable continuous prompts (encoding task-specific prior knowledge) and learnable conditional prompts (dynamically adapting to visual cues) to exploit CLIP's comprehensive semantic knowledge for discovering novel relationships. This multi-modal prompting strategy aligns visual-language representations through subject, object, union, background, and motion information representations. Secondly, we design a dynamic prompt learning approach to transfer the open-vocabulary knowledge into our model for predicting final relation categories. Extensive experiments on the public VidVRD detection benchmark datasets demonstrate that our method shows significant performance improvements over existing approaches, by an improvement of 28.86$\%$ mAP for all relation categories and an increase of 47.73$\%$  for unseen relation categories. 
In summary, our contributions in this paper are as follows:
\begin{itemize}
\item We introduce a new open-vocabulary VidVRD framework that incorporates automated region-caption generation for the first time. This approach improves VidVRD performance while also achieving state-of-the-art results.
\item Our proposed aggregation module is created to align visually with open-vocabulary text descriptions, greatly improving the representation of semantic information. This module enhances the model's capability to accurately grasp the connections between objects by integrating cross-modal spatiotemporal information, leading to a more refined semantic comprehension in challenging situations.
\item By incorporating semantic-alignment prompt learning, our model can improve the generalization capability by dynamically fusing learnable and hand-crafted prompts.
\end{itemize}
\section{Related Works}
\subsection{Video Visual Relation Detection}
VidVRD\cite{cao2021vsrn,shang2017video,su2020video,zheng2022vrdformer,woo2021and,ni2023human,zhang2024entity} is focused on tracking relations between observed objects in videos using the format $\langle$``subject", ``predicate", ``object"$\rangle$. The field of research has gained significant momentum with the unveiling of expansive datasets such as ImageNet-VidVRD and VidOR. Shang et al.\cite{shang2017video} were the first to present the ImageNet-VidVRD dataset and also put forward a commonly utilized three-stage detection framework, which incorporated enhanced dense trajectory visual features, relative spatial positions, and semantic categorizations. Inspired by VidVRD, Su et al.\cite{su2020video} proposed a Multiple Hypothesis Association (MHA) method, to maintain various relation hypotheses and address inaccuracies or missing data, ultimately generating more precise relations. The Transformer-based framework VRDFormer\cite{zheng2022vrdformer} combined stages in an autoregressive manner, created relational instances through query methods and utilized static and circular queries to monitor object pairs in a spatiotemporal context. Furthermore, VrdONE\cite{jiang2024vrdone} integrated relation classification and relation mask generation per frame into a single step, effectively capturing both short and long-term relationships by enhancing the interaction between subject and object in both temporal and spatial dimensions. Although current VidVRD methods have shown impressive results, they are restricted to closed-set scenarios where the training and testing datasets contain the same object and relation categories. As a result, they encounter difficulties in adjusting to the changing environments in real-life video situations. EDLN\cite{zhang2024entity} captures dependency information between relation predicates and their corresponding subject, object, and subject-object pairs,
\subsection{Open-Vocabulary Video Visual Relation Detection}
Annotations on datasets often show an imbalance between relation triples and object labels, resulting in a skewed distribution. The bias is further highlighted in videos due to the inclusion of dynamic relations in the temporal dimension\cite{cao2022concept}. CKERN\cite{cao2023video} began by considering basic characteristics and prior language connections, viewing identified pairs of objects in relationships as inquiries to gather relevant information from a knowledge base and structuring them in a coherent way to establish the semantic context of the relationships. However, predefined information often restricts the scope of common sense knowledge bases, hindering the possibility of achieving true open-vocabulary VidVRD. RePro\cite{gao2023compositional}  was recognized as the first pioneer to deal with open-vocabulary VidVRD. Wang et al.\cite{yang2024multi} further developed this idea by employing multi-modal prompts to uncover new relations. In contrast to them, we emphasize semantic space alignment between visual regions and texts, dynamically providing prompts for relation prediction based on various visual regions.\vspace{-1pt}
\subsection{Prompt Learning}
Pre-trained vision-language models (VLMs)\cite{radford2021learning,li2022blip,dong2019unified,luo2020univl,wang2024pedestrian} have achieved remarkable advancements across various downstream visual-language tasks.  Different prompt learning strategies have been applied to adjust VLMs for subsequent tasks. In Context Optimization\cite{zhou2022learning}, the hand-crafted prompts were substituted with learnable prompt tokens. kgCoOp\cite{yao2023visual} added a context loss function guided by knowledge to improve the model's capacity for generalization to new categories. CoCoOp\cite{zhou2022conditional} presented Meta-Net, a lightweight network that produced an input-conditional token for each image. This token was then integrated with a learned context vector, enabling the model to learn broader prompts. However, relying solely on quick learning in either language or visual aspects of CLIP is not ideal. Thus, multi-modal prompt learning, as demonstrated by recent studies\cite{khattak2023maple,gao2024clip,zhu2023prompt,yao2024tcp,yin2024hierarchy} had gained popularity for its incorporation of cross-modal knowledge, thereby improving the flexibility of multi-task visual language learning. For instance, MaPLe\cite{khattak2023maple}, conducted prompt learning on language and vision branches. These methods are specifically created for static images. On the other hand, we have created a new learning method that is designed specifically for open-vocabulary VidVRD and is effective.

\section{Method}
\subsection{Preliminary}
\textbf{Problem Setup.} The objective of VidVRD is to identify a series of visual relation triples and the motion trajectories of entities within a video \(V\). The final visual relation instance is represented as a tuple \((\langle s, r, o \rangle, T_{t_1:t_n}^s, T_{t_1:t_n}^o))\), where s, r, and o represent categories of subjects, relations, and objects, respectively. The trajectories  \(T_{t_1:t_n}^s=(b_{t_1}^s, \ldots, b_{t_n}^s) \), \(T_{t_1:t_n}^o=(b_{t_1}^o, \ldots, b_{t_n}^o)\) represent 
 temporally continuous bounding box sequences for the subject and object, with \(b_{t_i}^s\) and \(b_{t_i}^o\) specifying their spatial positions at each timestamp  \( T_{t_1:t_n}^{s} \). In open-vocabulary VidVRD, the dataset categories are divided into base and novel categories, with object categories represented as \( O_{base} \) and \(O_{novel} \), and relation categories represented as \( R_{base} \) and \( R_{novel} \), respectively. We train our model using visual relational triples that include base categories, while inference is performed on both novel and all categories.

\begin{figure*}[!t]
    \centering
    \includegraphics[width=\textwidth]{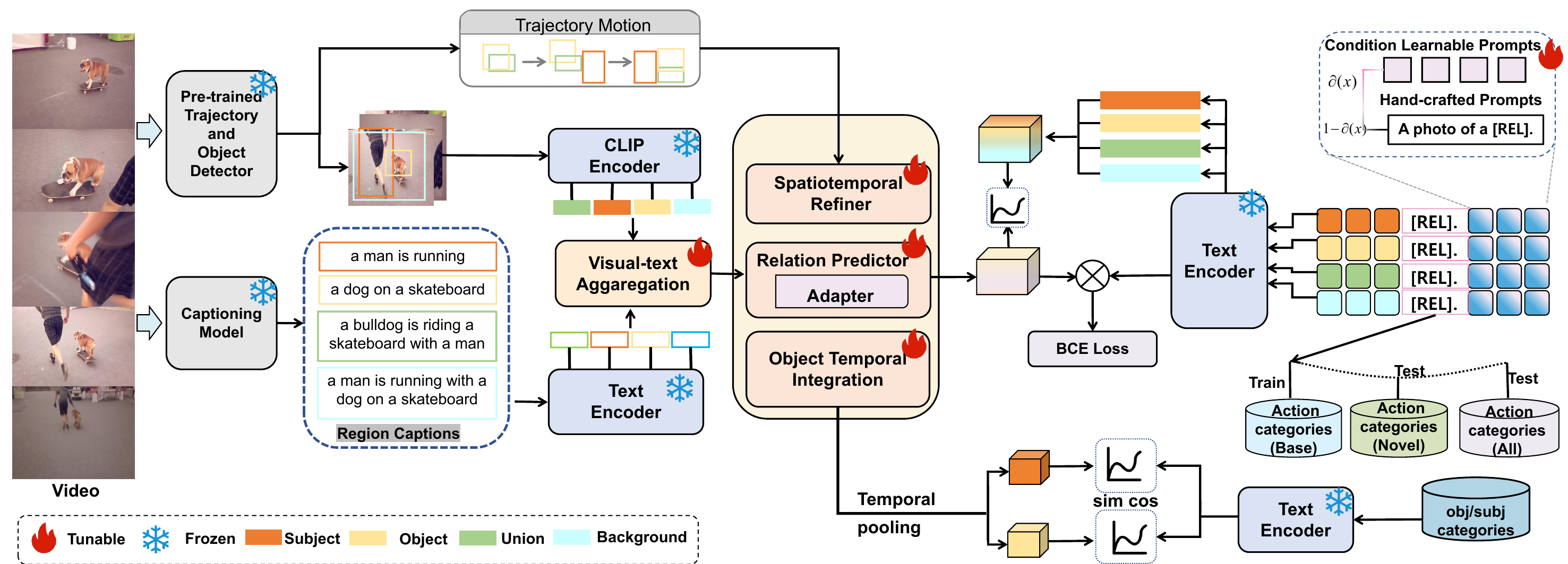}
    \caption{Overview of our framework.  The proposed framework breaks down VidVRD into two main tasks: detecting object trajectories and classifying relations. Initially, trajectories for all objects are extracted from the video, followed by the generation of region captions from the caption model. Afterward, a module for aggregating visual and text information is created to combine data from different modes. To capture both spatial and temporal contexts, we use a spatiotemporal refiner that takes in subject, object, union, background, and motion information in sequence. To examine how visual regions align with relationships, we combine both learnable and hand-crafted prompts to adjust a vision-language model for predicting new relations.}
    \vspace{-0.3cm}
    \label{framework}
\end{figure*} 

\noindent \textbf{Overall Architecture.} Figure~\ref{framework} depicts the framework we have put forward, intending to achieve open-vocabulary VidVRD by capturing the semantic connection between visual regions and relations. Our proposal consists of four folds: an open-vocabulary object tracklet detection module, a visual-text aggregation module to improve video representation, a spatiotemporal refiner for modeling spatiotemporal relations, and a dynamic prompting module with both hand-crafted and condition-learnable prompts based on visual regions.

\subsection{Open-Vocabulary Tracklet Detection}
In open-vocabulary target tracking detection, the process is divided into two main stages: generating trajectory proposals and classifying trajectories. Initially, we employ a pre-trained trajectory detector\cite{gao2022classification} to extract trajectories for all objects present in the video, referred to as \({T}_i^{\text{s/o}} = \{b_i^{\text{s/o}} \}_{i=1}^N\), where each trajectory contains a sequence of bounding boxes with RoI-aligned spatiotemporal visual features. Additionally,  for every subject-object trajectory pair \((\langle T^s,T^o\rangle \), we extract motion patterns \(  m_{s,o}\) encoding their relative positional dynamics and motion trends across timestamps. These features are aggregated via a mapping function , to obtain the features of motion information for all subject-object pairs as: \( M_{\text{s,o}} = \phi_{\text{mot}}\left(\{m_{\text{s,o}}^{s,o}\}\right)\).We leverage CLIP to extract visual representations from trajectory-aligned regions, \textit{i.e.,} \( v_i = \phi_o(f_i) \), which facilitates classification for each trajectory. We extract the text embeddings \(t_c\)  for all categories of objects in advance. Next, we compare each tracklet with all text embeddings and assign the object category label that corresponds to the highest score. That is:
\begin{align}
p(c) &= \frac{\exp\left(\frac{\cos(v_i, t_c)}{\tau}\right)}{\sum_{c\in C_{\text{b}} \cup C_{\text{n}}} \exp\left(\frac{\cos(v_i, t_c)}{\tau}\right)}
\label{eq:sim}
\end{align}
where \( \tau \) is the temperature parameter of softmax, and \( \cos(v_i, t_c) \) indicates the cosine similarity between  the \( i \)-th object trajectory of the visual representation \( v_i \)  and the text embedding \( t_c \) corresponding to the object category \( c \).

\subsection{Visual-Text Aggregation}
We utilize Blip2\cite{li2022blip} to generate region captions, denoted as \( Cap_f \), where \( Cap_f = \{ Cap_f^1, Cap_f^2, \ldots, Cap_f^T \} \) and \( Cap_f^i \) represents the generated caption for frame \( f_i \). Then,  a visual-text aggregation module is created to improve video representation by including detailed information about subjects, objects, and their relations in region captions\cite{gan2022vision}. A text encoder from CLIP is utilized to encode each region's captions. The semantic descriptions are concatenated in the set \( S = \{ s^k \}_{i=1} \), $ S {\in}\mathbb{R}^{T \times 4 \times d}$, where d is the dimension of the text feature, \ \( k \in \{ s, o, u, b \} \) denotes the subject region (\( s \)), the object region (\( o \)), the union region covering both the subject and object (\( u \)), and the background representing the entire image (\( b \)). The subject region typically includes the object, and extracting its features is useful to identify the object's attributes, such as shape and color. The object region emphasizes secondary objects connected to the subject, helping to clarify their relations. The union region refers to the spatial connections between the subject and object, such as their position and distance from each other, which contributes to capturing how the objects interact. The background offers environmental details that help the viewer understand the context of the entire scene.

As shown in Figure~\ref{module}, we employ self-attention mechanisms and feedforward layers of the Transformer to better analyze text features over time, enhancing the overall comprehension of text representations. Afterward, the text features are used to spatially modify visual features and combine information from multiple modes using dot cross-attention. Finally, the resulting features undergo spatial decomposition using spatial global normalization, and a spatiotemporal transformer is employed to aggregate temporal correlations. The entire process is formulated as:
\begin{align}
    \tilde{f_{vt}} = VT(f_v, t)   
\end{align}
where VT represents the visual-text aggregation module,  $\tilde{f_{vt}} {\in}\mathbb{R}^{T \times 4 \times d}$ is a cross-modal feature of the output and contains textual semantic characteristics.

\subsection{Spatiotemporal Refiner Module}
\begin{figure}[]
    \centering
    \includegraphics[width=\linewidth]{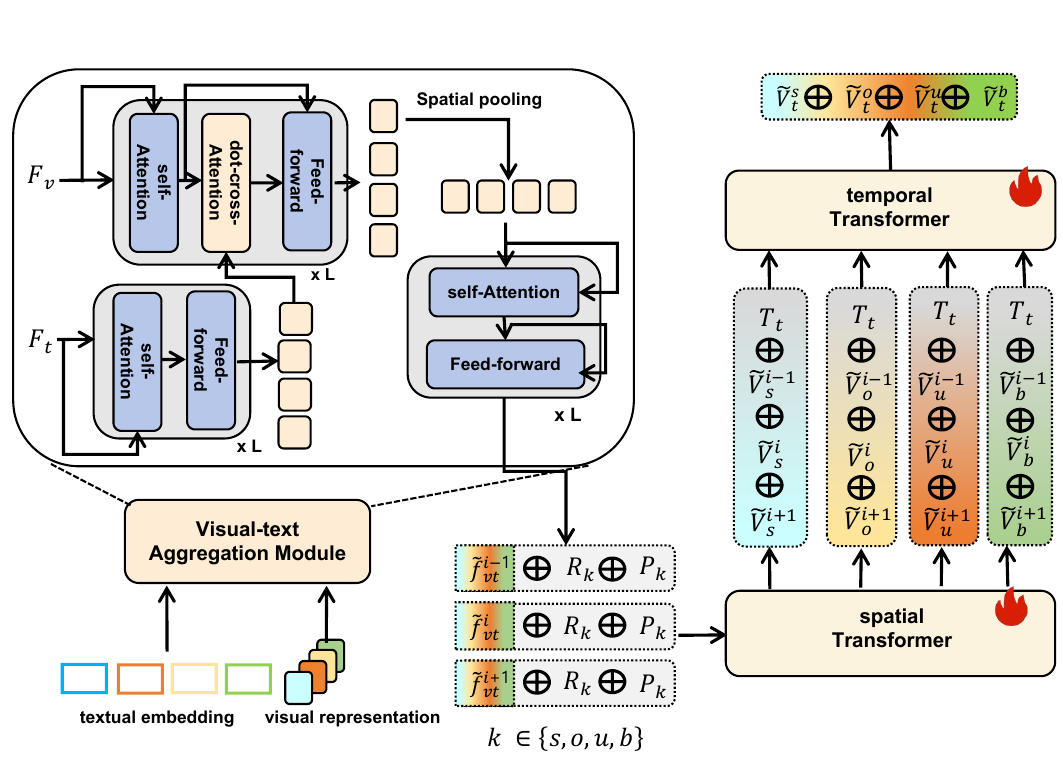}
    \caption{Overview of visual-text aggregation module and spatiotemporal refiner. The visual-text aggregation module combines visual and textual representations using self-attention and dot-cross attention. It further refines these features through feedforward networks and spatial pooling operations, allowing it to effectively capture the correlations between visual contents and region captions. Meanwhile, The spatiotemporal refiner module employs a spatiotemporal transformer to examine dynamic state variations, updating visual representations successfully.}
    \label{module}
\end{figure}

The spatiotemporal refiner module (STRM) inspired by STTran\cite{cong2021spatial}
is composed of a spatial Transformer and a temporal Transformer. The architecture is illustrated in Figure~\ref{module}. We connect information about subjects, objects, unions, backgrounds, and motion in a sequence to feed into the spatial Transformer to make our model recognize the visual features and motion trends based on spatial and semantic relations. To enhance the model's understanding of visual features in visual scenes, we incorporate role embedding \(R^k\),\ \( k \in \{ s, o, u, b \} \) to differentiate between the roles of subjects, objects, their union, and the background. Briefly, we formulate the visual features as:
\begin{align}
    \tilde{V^k} = STran(\tilde{f_{vt}} + R_k + P_k + M_{s,o}), k \in \{ s, o, u, b \}
\end{align}
where \(\text{STran}(\cdot)\) denotes the spatial Transformer and \(P_k\) are the positional embeddings.

The temporal Transformer can capture their dynamic changes and interactions throughout the entire time series. Specifically, we concat the features of each role across all frames $\tilde{V}$ in the temporal encoder, adding temporal embeddings \(T_t\) structured in a similar way to role embeddings. These are essential for detecting and predicting time-related object interactions, such as ``push" and ``pull". The multi-head self-attention layer extracts the temporal dependencies between contextual features, enabling the model to understand the evolution of these features over time. 
\begin{align}
    \tilde{V_t} = TTran(\tilde{v_{t}} + T_t )
\end{align}
where \(\tilde{V_t}\) = \(\tilde{V_t^s}\)+\(\tilde{V_t^o}\)+\(\tilde{V_t^u}\)+\(\tilde{V_t^b}\) and \(\text{TTran}(\cdot)\) stands for the temporal Transformer. 

\subsection{Prompt-Driven Semantic Alignment Learning}
\label{subsection:prompt}
\textbf{Open-Vocabulary Object Classification.} The visual representations of the spatiotemporal refiner module (STRM) are fed into an open-vocabulary object classifier, which generates object classification results by computing the similarity between visual and textual representations. The classifier employs CLIP to compute the similarity between these visual representations and text embeddings in Eq.\eqref{eq:sim}. To enhance the classification performance of novel object categories, we incorporate additional bottleneck layers in visual and text branches that facilitate learning new features and their integration with the original trained features through residual connections, thus can be represented by
\begin{align}
\text{Adapter}_{v}(F) = \text{BN}(\sigma(W[\tilde{V_t} + H]))
\end{align}
\vspace{-0.75cm}
\begin{align}
\text{Adapter}_{t}(F) = \text{BN}(\sigma(W[\tilde{W_t} + H]))
\end{align}
where \(\tilde{V_t}\) and \(\tilde{W_t}\) are obtained from the original STRM module and CLIP' text encoder, respectively, \(H\) denotes the features learned by the additional adapter layer, \(W\) is the weight of the adapter layer, \(\sigma\) is the nonlinear activation function, and \(BN\) indicates batch normalization.

\noindent \textbf{Open-Vocabulary Relation Classification.} Unlike traditional classification tasks, the number of categories to predict is unknown during both training and inference.  The open-vocabulary model relies on pre-extracted relations for predictions. Therefore, we introduce a dynamically prompt learning strategy into our framework to comprehend and recognize novel categories.
Regarding each relation category \([REL]\),\ \( [REL] \in \{C_b^r\} \) is developed for training while \([REL] \in C_b^r \cup C_n^r \) for testing. We introduce \(M\) learnable context prompts \(\{v_1, v_2, \cdots, v_M\}\), which share the same dimensionality as the text embeddings. The prompt for the \( i \)-th category, represented by \( \varsigma_i \), is defined as \( \varsigma_i = \{v_1, v_2, \ldots, v_M, c_i\} \), where \( c_i \) corresponds to the text embedding for the relation label. The initialization and training of \( N_\varsigma \)-token language prompts are dedicated to four distinct roles: subject, object, union, and background. Each set of prompts is defined as follows. 
\begin{align}
 \varsigma^k = [v_1^k, v_2^k, \ldots, v_M^k, c_i^k\}]
\end{align}
where \( k \) represents the specific role \( \{s, o, u, b\} \), and the learnable context prompts are shared across all categories.

The hand-crafted prompts can provide a more foundational understanding, as opposed to learnable prompts, which may be more specialized or context-dependent. Therefore, we combine the learnable prompts with the manual prompts in a dynamic manner. Let \(T(\cdot)\) denote the text encoder of CLIP, and the final text features of relation category $r$ are defined by
\begin{align}
    text_i^r  = \alpha(x) \varsigma_i^k + (1 - \alpha(x)) t_{i^r,\text{hand}},
\end{align}
\vspace{-0.75cm}
\begin{align}
         t_r = T(text_i^r)
\end{align}
\subsection{Training Objective}
To detect the open-vocabulary VidVRD, our training loss consists of three parts: a relation contrastive loss \(L_{rel}\), an object and subject contrastive loss \(L_{obj\&sub}\), and an interaction loss \(L_{int}\).  The overall training
loss is:
\begin{align}
    L =  \gamma L_{obj\&sub} + L_{\text{rel}} + \delta L_{\text{int}}
\end{align}
where \(\gamma\) and \(\delta\) are hyperparameters.

\noindent \textbf{Object and Subject Contrastive Loss.}  We employ an object and subject contrastive loss to prevent visual representations from drifting due to prompt-driven semantic alignment learning. This loss guarantees that the characteristics obtained from the spatiotemporal transformer are able to effectively differentiate among various objects. Specifically, all subject and object features of frames are average-pooling after spatiotemporal modeling, expressed as \(\tilde{v}_k = \text{TimeAvg} \left( \{\dot{v}_t^k \}_{t=0}^{T} \right),\quad k \in \{s, o\}\). Meanwhile, We obtain the text embeddings for all subject or object categories by hand-crafted prompts (e.g., ``a photo of [CLS]") into the text encoder of CLIP. The similarity between the visual and language representations of the category is typically calculated by
\begin{align}
    \hat{y}^{k} = \cos(\bar{v}^{k}, \hat{l}^{c}), \quad k \in \{s, o\}.
\end{align}
Hence, the object and subject contrastive loss is computed using the cross-entropy loss (CE):
\begin{align}
    L_{obj\&sub} =  \text{CE}(\hat{y}^{o}, y^{o})+\text{CE}(\hat{y}^{s}, y^{s}) 
\end{align}
where \(\hat{y}_s\) represents the predicted subject similarity between the visual and language representations, while \(\hat{y}_o\) denotes the corresponding predicted object similarity. Additionally, \(y_s\) and \(y_o\) represent the ground-truth category labels for the subject and object, respectively.

\noindent \textbf{Relation Contrastive Loss.} The relation category prediction score \(c_r\) is determined by
\begin{align}
    \hat{y}_r^{\text{rel}} = \sigma\left(\cos\left(\tilde{v}, \tilde{l}_r\right)\right)
\end{align}
where \(\tilde{v}\) represents the visual representations, \(\tilde{v} = \cup \{ \ddot{v}_s, \ddot{v}_o, \ddot{v}_u, \ddot{v}_b \}\); \(\tilde{l}_r\) represents the text representations, \(\tilde{l}_r = \cup \{ \ddot{l}_r^s, \ddot{l}_r^o, \ddot{l}_r^u, \ddot{l}_r^b \}\); \(\sigma()\) is the sigmoid function; \(cos()\) is the cosine similarity. The binary cross-entropy (BCE) is designed for loss calculation because it calculates the loss for each relation category separately without considering other categories. Thus, we get
\begin{align}
    L_{\text{rel}} = \frac{1}{|C^R_b|} \cdot \sum_{c \in C^R_b} \text{BCE}(\hat{y}^{\text{rel}}_c, y^{\text{rel}}_c)
\end{align}
where \(y^{\text{rel}}_c\) is set to \(1\) if and only if the category is the ground-truth predicate category, otherwise, \(y^{\text{rel}}_c\) is set to \(0\).

\noindent \textbf{Interaction Loss.} We assess every combination of subjects and objects within the frames, even if they show no interaction. If there is a relation between them, we assign their interaction by \(y^{int}=1\); Conversely, if there is no relation, we assign by \(y^{int}=0\). Thus, the interaction loss is computed using the binary cross-entropy loss as:
\begin{align}
L_{\text{int}} = \frac{1}{T_e-T_s} \sum_{t=T_s}^{T_e} \text{BCE}(\hat{y}_{\text{int}}, y_{\text{int}})
\end{align}
where all features in the frame are concatenated to predict the interaction probability, represented as \(\hat{y}_{\text{int}} = \cup \{ \ddot{v}_s, \ddot{v}_o, \ddot{v}_u, \ddot{v}_b \}\). \(T_s\) and \(T_e\) are the start and end time of the
frame, respectively.

\section{Experiments}
\subsection{Datasets and Evaluation Metrics.} 
\begin{table*}[t]
\caption{\textbf{Comparison with SOTA open-vocabulary VidVRD methods on VidVRD dataset.} The experiments are conducted under the All-split and Novel-split settings.}
\centering
\begin{tabular}{c|c|ccc|ccc|ccc}
    \toprule
    \multicolumn{1}{c|}{\multirow{2}{*}{Splits}} &\multirow{2}{*}{Model} & \multicolumn{3}{c|}{SGDet}  & \multicolumn{3}{c|}{SGCls} & \multicolumn{3}{c}{PredCls}\\
    \cline{3-5}
    \cline{6-8} 
    \cline{9-11}
    & & mAP & R@50 & R@100 & mAP & R@50 & R@100 & mAP & R@50 & R@100  \\
    \hline
    \multirow{5}{*}{All}  & ALPro\cite{he2022towards} & 3.03 & 2.57 & 3.11 & 3.92 & 3.88 & 4.75 & 4.97 & 4.50 & 5.79 \\
    \multirow{4}{*}{} & VidVRD-II\cite{shang2021video} & 12.66 & 9.72 & 12.50 & 17.26 & 14.93 & 19.68 & 4.97 & 4.50 & 5.79 \\
    \multirow{4}{*}{}     & RePro\cite{gao2023compositional} & 21.12 & 12.63 & 15.42 & 30.15 & 19.75 & 25.00 & 34.90 & 25.50 & 32.49 \\
    \multirow{4}{*}{}  & MMP\cite{yang2024multi} & 22.10 & 13.26 & 16.08 & 29.38 & 23.56 & 28.89 & 38.08 & 30.47 & 37.46 \\
    \multirow{4}{*}{}   & \textbf{Ours} & \textbf{28.48} & \textbf{17.15} & \textbf{19.90} & \textbf{30.25} & \textbf{23.38} & \textbf{30.66} & \textbf{38.64} & \textbf{31.56} & \textbf{38.42} \\
    \hline
    \hline
    \multirow{5}{*}{Novel}  & ALPro\cite{he2022towards} & 0.98 & 2.79 & 4.33 & 3.69 & 7.27 & 8.92 & 4.09 & 9.42 & 10.41 \\
    \multirow{4}{*}{}  & VidVRD-II\cite{shang2021video} & 3.11 & 7.93 & 11.38 & 5.70 & 13.22 & 18.34 & 7.35 & 18.84 & 26.44 \\
    \multirow{4}{*}{}   & RePro\cite{gao2023compositional} & 5.87 & 12.75 & 16.23 & 10.32 & 19.17 & 25.28 & 12.74 & 25.12 & 33.88 \\
    \multirow{4}{*}{}   & MMP\cite{yang2024multi} & 12.15 & 13.75 & 15.21 & 17.57 & 21.98 & 28.43 & 21.14 & 30.41 & 37.85 \\
    \multirow{4}{*}{}   & \textbf{Ours} & \textbf{17.95} & \textbf{17.69} & \textbf{19.01} & \textbf{18.25} & \textbf{28.98} & \textbf{31.85} & \textbf{21.96} & \textbf{31.24} & \textbf{39.08} \\
     \bottomrule
    \end{tabular}
    \label{vidvrd}
\end{table*}
\noindent\textbf{Datasets.} The VidVRD\cite{shang2017video} dataset contains 1,000 videos, 800 for training and 200 for testing. VidVRD covers 35 object classes and 132 predicate classes. In contrast, the VidOR\cite{shang2019annotating} dataset is even larger, containing 10,000 videos, 7,000 for training, 835 for validation, and 2,165 for testing. VidOR covers 80 object categories and 50 predicate categories, and the average length of the video is 34.6 seconds, with a total duration of approximately 98.6 hours. This dataset provides intensive annotations for each frame, including physical bounding boxes and detailed time boundaries for visual relations between subjects and objects.

\noindent\textbf{Metrics.} 
We evaluate our framework on the three standard VidVRD tasks, as delineated by Motifs\cite{zellers2018neural}: scene graph detection (SGDet), scene graph classification (SGCls), and predicate classification (PredCls). In SGDet, we identify object trajectories from raw video data and classify the relations among these objects. SGCls entails classifying the objects within the provided ground-truth trajectories while predicting their interrelations. PredCls is centered on predicting the relations between known objects, leveraging both the ground-truth trajectories and their corresponding categories. Two primary metrics we use are mean Average Precision (mAP) and Recall@K (R@K, K=50, 100). Specifically, a detected relation triplet is considered correct when it matches a corresponding triplet in the ground truth, and the IoU between the trajectories is greater than a threshold of 0.5.  

\noindent\textbf{Evaluation Settings.} For the open-vocabulary evaluation, the base and novel categories are determined by their frequency of occurrence.  We select common object and predicate categories as base categories, while rare ones are designated as novel categories. Our model is trained on base categories. The testing process follows two settings: (1) All-split evaluation considers all object and predicate categories.  (2) Novel-split evaluation includes all object categories and novel predicate categories.  Both evaluations are conducted on the test set of VidVRD and the validation set of VidOR, as the annotations for the VidOR test set are unavailable.
\subsection{Implementation Details}
Each video is sampled with 30 frames in all experiments.   To capture dynamic object instances across frames, we first employ the MEGA framework\cite{chen2020memory} initialized with ResNet-50\cite{he2016deep} parameters for per-frame detection.  These discrete detections are subsequently linked into continuous spatiotemporal pathways using the DeepSORT tracking algorithm\cite{wojke2017simple}, forming coherent object tracklets that serve as the foundation for feature extraction. Besides, the CLIP variant based on ViT-B/16  with frozen parameter weights is adopted. The multi-head self-attention in the spatiotemporal refiner blocks is configured with 8 heads, and the dropout rate is set to 0.1. Learning promptings are assigned 8 tokens, with the [CLS] token located at \(75\%\) of the total token length. Our model adopts the AdamW optimizer\cite{loshchilov2017decoupled} with an initial learning rate of \(10^{-3}\), applying a multi-step decay schedule at epochs 15, 20, and 25, during which the learning rate is reduced by a factor of 0.1 at each step. All experiments are conducted with a batch size of 32 on a single NVIDIA GeForce RTX 4090 GPU card.
\begin{table}[t]
  \caption{\textbf{Comparison with SOTA open-vocabulary VidVRD methods on VidOR dataset.} The experiments are conducted under the All-split and Novel-split settings. ``-" denotes the results are not available on the original papers.}
  \centering
  \begin{tabular}{c|c|ccc}
    \hline
    \multicolumn{1}{c|}{\multirow{2}{*}{Splits}} 
    &\multirow{2}{*}{Model}& \multicolumn{3}{c}{SGDet}  \\ 
    \cline{3-5}
    & & mAP & R@50 & R@100 \\
    \hline
    \multirow{2}{*}{All} 
    & MMP\cite{yang2024multi} & 7.15 & 6.54 & 8.29  \\
    \multirow{2}{*}{}  & \textbf{Ours} & \textbf{10.18} & \textbf{7.65} & \textbf{9.82}  \\
    \hline
    \hline
    \multirow{2}{*}{Novel} 
    & MMP\cite{yang2024multi} & 0.84 & 1.44 & 1.44  \\
    \multirow{2}{*}{} & \textbf{Ours} & \textbf{1.45} & \textbf{5.32} & \textbf{4.68} \\
    \hline
    \addlinespace
    \hline
   \multicolumn{1}{c|}{\multirow{2}{*}{Splits}} 
    &\multirow{2}{*}{Model}& \multicolumn{3}{c}{PredCls}  \\ 
   \cline{3-5}
    & & mAP & R@50 & R@100 \\
    \hline
    \multirow{5}{*}{All} & ALPro\cite{he2022towards} & - & 2.61 & 3.66  \\
    \multirow{5}{*}{}& CLIP\cite{radford2021learning} &1.29  &1.71 &3.13 \\
    \multirow{5}{*}{}& VidVRD-II\cite{shang2021video} & - & 24.81 & 34.11   \\
    \multirow{5}{*}{}& RePro\cite{gao2023compositional} & - & 27.11 & 35.76   \\
    \multirow{5}{*}{}& MMP\cite{yang2024multi} & 38.52 & 33.44 & 43.80   \\
    & \textbf{Ours} & \textbf{38.94} & \textbf{34.68} & \textbf{43.85}  \\
    \hline
    \hline
    \multirow{5}{*}{Novel} & ALPro\cite{he2022towards} & - & 5.35 & 9.79 \\
     \multirow{5}{*}{}&  CLIP\cite{radford2021learning} & 1.08 & 5.48 & 7.20 \\
    \multirow{5}{*}{} & VidVRD-II\cite{shang2021video} & - & 4.32 & 4.89 \\
    \multirow{5}{*}{} & RePro\cite{gao2023compositional} & - & 7.20 & 8.35  \\
    \multirow{5}{*}{} & MMP\cite{yang2024multi} & 3.58 & 9.22 & 11.53  \\
    \multirow{5}{*}{} & \textbf{Ours} & \textbf{3.87} & \textbf{10.24} & \textbf{12.56} \\
    \hline
    \end{tabular}
    \label{vidor}
\end{table}
\subsection{Comparison with State-of-the-art Methods}
We evaluate our framework against SOTA techniques on the VidOR and ImageNet-VidVRD benchmarks. Tables~\ref{vidvrd} and \ref{vidor} present the comparative results for the VidVRD and VidOR datasets, respectively. In conclusion, our model demonstrates competitive outcomes on both ImageNet-VidVRD and VidOR benchmarks. The findings from our comparative analysis of the VidVRD dataset are detailed in Table~\ref{vidvrd}.
The comparative results on the VidVRD dataset show that our method achieves the best performance in all evaluation metrics. 

Our proposal shows significant performance improvements in different scenarios, as shown in Table \ref{vidvrd}. Specifically, under the All-split setting, our model achieves an mAP of 28.48, which surpasses RePro\cite{gao2023compositional} by 28.87$\%$ and shows a notable improvement of 6.25 mAP over the model  in MMP\cite{yang2024multi}. Furthermore, in the PredCls task, our model achieves an mAP of 38.64, outperforming both RePro\cite{gao2023compositional}  and VidVRD-II\cite{shang2021video} by 10.72$\%$ and 67.75$\%$, respectively. In the Novel-split setting, similar results are obtained, where ours performs 5.8 mAP better than MMP. These results highlight how effective our approach is in improving the accuracy and reliability of Open-VidVRD tasks.

As presented in Table \ref{vidor}, our method demonstrates a slight improvement over other approaches, with a 16.98$\%$ increase in R@50 and an 18.46$\%$ increase in R@100 in the Novel-split setting on the SGDet task. Additionally, in the All-split setting, our method attains a more significant enhancement, with a 0.11$\%$ increase in R@100 and a 3.7$\%$ increase in R@50.
Notably, the VidOR dataset exhibits longer video durations compared to the ImageNet-VidVRD dataset, which frequently leads to diminished performance across identical metrics.

\subsection{Abaltion Studies}
We conduct extensive ablation studies on the benchmark dataset ImageNet-VidVRD to evaluate each component of our method in depth.These components work cooperatively to achieve robust and accurate open-vidvrd, demonstrating the effectiveness of the proposed framework.

\noindent\textbf{Effectiveness of different components.} To investigate the effectiveness of each component of our model, we conduct ablation studies on VidVRD dataset. Table \ref{components} presents the performance of each
model variant. 
\begin{table}[t]
      \caption{\textbf{Performance (mAP) of ablation study for different components on the VidVRD dataset.}  Note that we add linear layers to keep similar amount parameters when a module is absent. }
    \centering
    \begin{tabular}{c|cc|cc}
        \hline
        \multicolumn{1}{c|}{\multirow{2}{*}{Components}}
        & \multicolumn{2}{c|}{All} & \multicolumn{2}{c}{Novel} \\ 
        \cline{2-5}
        & SGDet & PredCls & SGDet & PredCls \\ 
        \hline
        \multirow{1}{*}{+ visual-text aggregated} & 13.56 & 31.89 & 9.69 & 13.74 \\
        \multirow{1}{*}{+ spatial transformer} & 21.16 & 35.58 & 15.09 & 19.62 \\
        \multirow{1}{*}{+ prompt} & \textbf{28.48} & \textbf{38.64} & \textbf{17.95} & \textbf{21.96} \\
        \hline
    \end{tabular}
    \label{components}
\end{table}

\noindent\textbf{Effectiveness of different aggregation manners.} We propose spatially fusing the text and visual features through the dot cross-attention operator in the visual-text aggregation module. Table \ref{manner} compares the different aggregation methods. As shown in Table \ref{manner}, the term ``Concat/Sum" refers to connecting/summing the visual features from the spatial transformer and the text description from the text encoder of CLIP before inputting them into the temporal transformer for multimodal fusion. Our module design's high efficiency is confirmed by the fact that the cross-attention variant outperforms all others in every scenario.
\begin{table}[t]
\caption{\textbf{Ablation study for different aggregation manners on the VidVRD dataset.}}
    \centering
    \begin{tabular}{c|cc|cc}
        \hline
        \multicolumn{1}{c|}{\multirow{2}{*}{Aggregation}}
        & \multicolumn{2}{c|}{All} & \multicolumn{2}{c}{Novel} \\ 
        \cline{2-5}
        & SGDet & PredCls & SGDet & PredCls \\ 
        \hline
        \multirow{1}{*}{Sum} & 22.34 & 34.26 & 15.78 & 18.65   \\ 
        \multirow{1}{*}{Concat} & 19.56 & 33.75 & 14.34 & 19.89 \\
        \multirow{1}{*}{Cross-Attention} & \textbf{28.48} & \textbf{38.64} & \textbf{17.95} & \textbf{21.96} \\
        \hline
    \end{tabular}
    \label{manner}
\end{table}
\begin{table}[t]
     \caption{\textbf{Ablation study for three variants of adapters on the VidVRD dataset.} ``Vis" represents the visual adapter, ``Txt" denotes the text adapter.}
    \centering
    \begin{tabular}{c|c|cc|cc}
        \hline
        \multicolumn{1}{c|}{\multirow{2}{*}{Vis}} 
        & \multicolumn{1}{c|}{\multirow{2}{*}{Txt}}
        & \multicolumn{2}{c|}{All} & \multicolumn{2}{c}{Novel} \\ 
        \cline{3-6}
        & & SGDet & PredCls & SGDet & PredCls \\ 
        \hline
        \multirow{1}{*}{\checkmark} & \checkmark & 27.88 & 37.52 & 17.09 & 21.50 \\
        \multirow{1}{*}{-} & \checkmark & 28.27 & 37.93 & 17.05 & 21.58 \\
        \multirow{1}{*}{\checkmark} & - & \textbf{28.48} & \textbf{38.64} & \textbf{17.95} & \textbf{21.96}  \\
        \hline
    \end{tabular}
    \label{adapter}
    \vspace{-0.3cm}
\end{table}
\begin{table}[t]
\caption{\textbf{Ablation study for different prompt learning variants on the VidVRD dataset.}}
    \centering
    \begin{tabular}{c|cc|cc}
        \hline
        \multicolumn{1}{c|}{\multirow{2}{*}{Prompt Variants}}
        & \multicolumn{2}{c|}{All} & \multicolumn{2}{c}{Novel} \\ 
        \cline{2-5}
        & SGDet & PredCls & SGDet & PredCls \\ 
        \hline
        \multirow{1}{*}{Hand-crafted}  &27.35 &37.10  &13.04 &17.83 \\ 
        \multirow{1}{*}{Continuous} &28.27 &37.94 & 15.21 &19.21 \\
        \multirow{1}{*}{Conditional} & 26.85 & 37.44 & 17.29 & 19.88  \\
        \multirow{1}{*}{Our} & \textbf{28.48} & \textbf{38.64} & \textbf{17.95} & \textbf{21.96} \\
        \hline
    \end{tabular}
    \label{prompt}
\end{table}
\begin{figure}[!ht]
    \centering
     \includegraphics[width=0.9\linewidth,height=5cm]{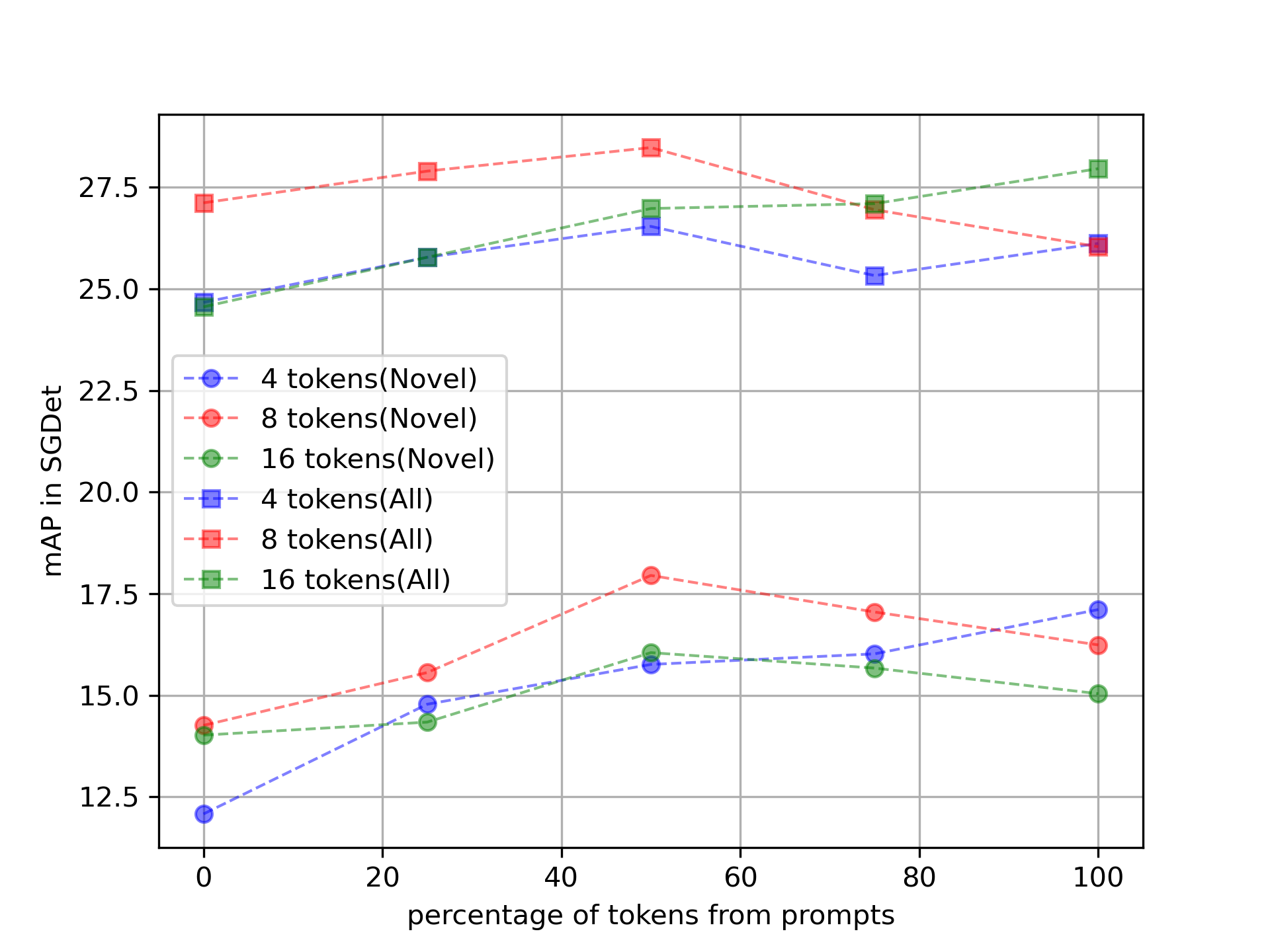}
     \caption{\textbf{Ablation study for prompt lengths.} The results illustrate how different prompt lengths impact the outcomes, using blue for 4 tokens, red for 8 tokens, and green for 16 tokens. The horizontal axis represents the token percentage, from 0$\%$  (all tokens from learnable prompts) to 100$\%$  (all from hand-crafted prompts).}
    \label{prompt_length}
    \vspace{-0.3cm}
\end{figure} 
\begin{figure*}[]
 \caption{\textbf{Qualitative examples of our model}. The novel relations detected by the model are marked in yellow background.}
    \centering
     \includegraphics[width=1\textwidth]{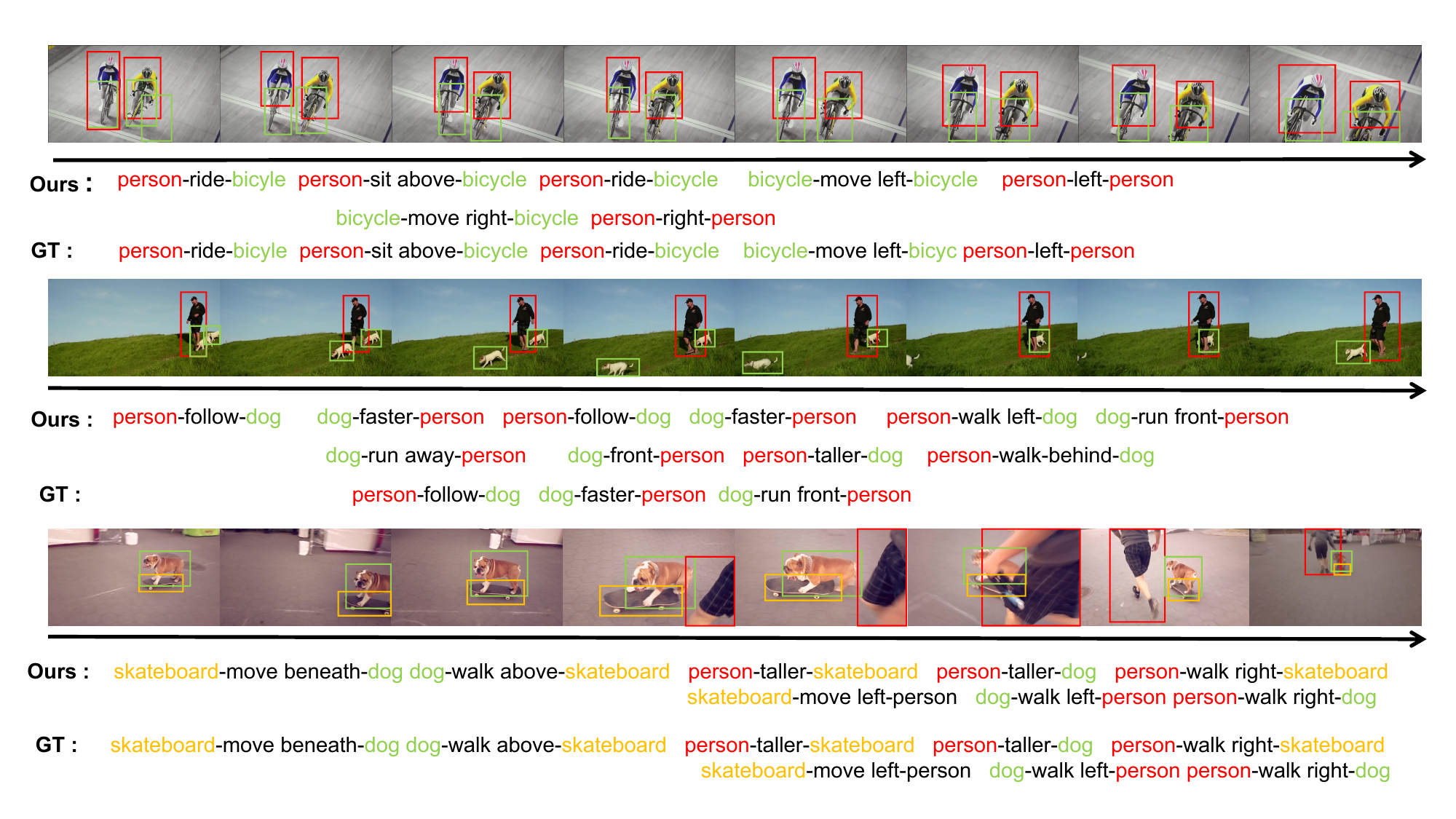}
    \vspace{-1cm}
\end{figure*} 
\begin{figure*}
    \centering
     \includegraphics[width=1\textwidth]{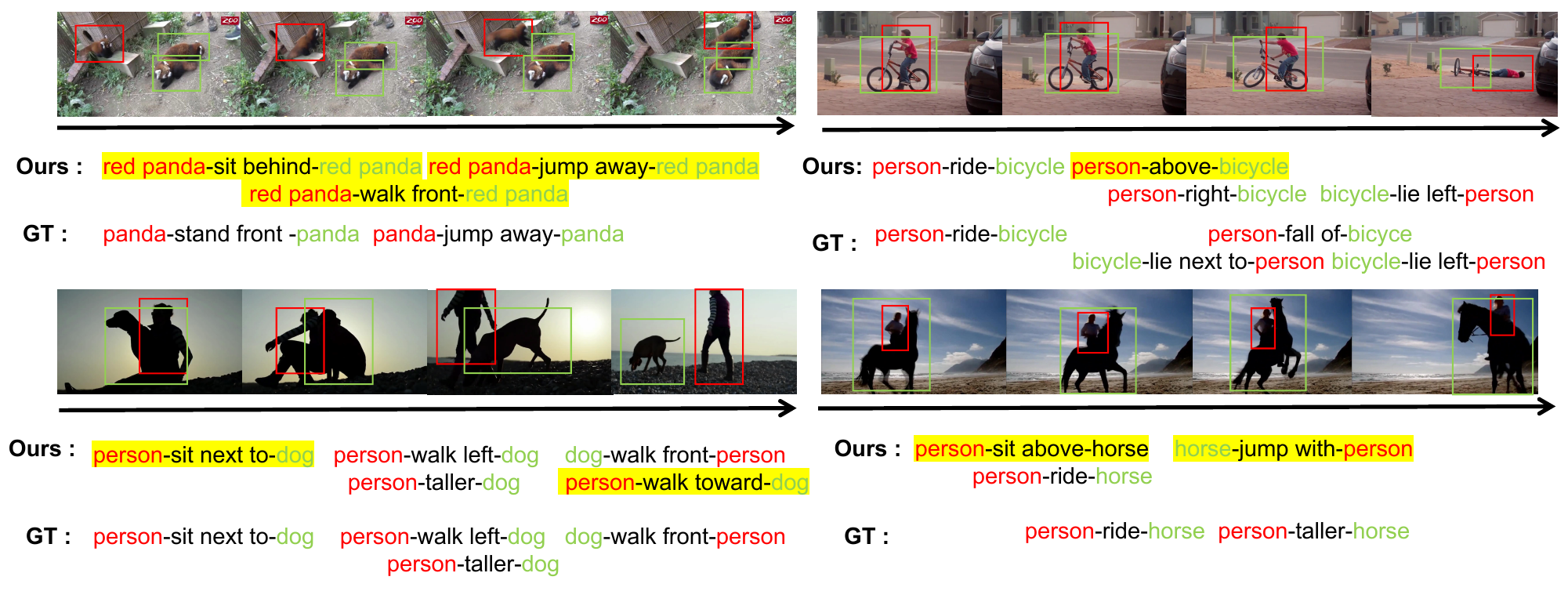}
    \label{visual}
    \vspace{-0.3cm}
\end{figure*}

\noindent\textbf{Effectiveness of different adapter variants.} As mentioned in Section \ref{subsection:prompt}, two variants of adapter have been explored. The first strategy involves fine-tuning the text adapter while keeping the visual adapter fixed. The second strategy entails fine-tuning both the text and visual adapters simultaneously. Our experiment shows that both adapters performed at similar levels when compared, as illustrated in Table \ref{adapter}. Notably, the visual adapter performed better than the text adapter when used alone. The variance could be attributed to the visual adapter's capacity to offer a more direct and enhanced depiction of characteristics in various visual activities. Furthermore, although there is some performance improvement when combining the text and visual adapters, it does not outperform the effectiveness of using just the visual adapter. This indicates that the two adapters may duplicate the information they collect, which will cause redundancy or interference.

\noindent \textbf{Effectiveness of different prompt learning variants.} To assess how well our prompting learning works, we create three different versions of our method to compare. (1) The ``Hand-crafted" approach employs predefined templates to address various narrative elements: for subjects, it uses templates like ``An image of a person or object [CLS] something"; for objects, it uses templates like ``An image of something [CLS] a person or object"; and for unions and backgrounds, it applies templates like ``An image of the visual relation [CLS] between two entities". (2) ``Continuous" is a unified learning vector based on continuous prompt learning; (3) "Conditional" involves generating all continuous prompts for input visual features. From the results shown in Table \ref{prompt}, it is clear that our prompt learning is superior to other variants. Specifically, we achieved a substantial gain of nearly 5$\%$ in the novel split setting. Furthermore, our results indicate that learnable prompts (both continuous and conditional) consistently outperform the Hand-crafted ones.

\noindent \textbf{Effectiveness of different prompt lengths.} We conduct experiments using 4, 8, and 16 tokens at different percentages to examine how varying token numbers influence performance. The analysis presented in Figure~\ref{prompt_length} reveals that increasing the number of tokens initially enhances performance, followed by a decline. Moreover, as the proportion of markers suggested by learnable conditions rises, the results show initial improvement but then exhibit instability. The best performance is achieved when the token count reaches 16, with half of the tokens coming from prompts that can be learned. These findings highlight how important it is to blend task-specific expertise with prompts and validate the benefit of using both manual and learnable prompts together.
\subsection{Qualitative Analysis}
Several visualization examples are presented to showcase the strengths of our model. For each video, we have selected keyframes that effectively represent the video's context. Ground truth triplets (GT) are displayed to enhance the visualizations' intuitiveness. Our method efficiently detects object trajectories and accurately classifies relations between objects in both base and novel categories. It effectively addresses the difficulties of identifying objects that are only partially visible and intricate, dynamic relationships like ``jump away". In complex environments, as illustrated in figure~\ref{visual}, our model reliably detects object trajectories and interaction classes, demonstrating its robustness across various scenarios. However, we observe in our visualization findings that despite thorough annotation, certain relationships are still absent. For example, although our model correctly identifies the $\langle$ person, walk toward, dog $\rangle$, this relation does not exist in the annotations, suggesting that omissions in relation labeling are difficult to avoid. This underscores the challenges involved in relational annotation, emphasizing the importance of open-vocabulary detection.

\section{Conclusion}
In this work, we propose a prompt-driven semantic alignment (OpenVidVRD) framework for open-vocabulary VidVRD. Our proposed OpenVidVRD seamlessly incorporates spatiotemporal information into VLMs and ensures satisfactory semantic space alignment between visual regions and text representations, overcoming the constraints of conventional VidVRD approaches. Furthermore, we utilize a proactive prompting approach to improve the model's effectiveness in classifying relations. Extensive experiments on public datasets confirm the effectiveness and reliability of the proposed method, showing significant enhancement in detecting new relations while still performing well in base categories. Our goal moving forward is to create a comprehensive model that can detect both object trajectories and relations at the same time, improving our comprehension of object interactions in real-world scenarios, especially in live videos.

\section{Declaration of competing interest}
The authors affirm that they possess no recognized competing financial interests or personal relationships that may have the potential to influence the findings presented in this paper.

\bibliographystyle{IEEEtran}

\bibliography{reference}

\newpage
\begin{IEEEbiography}[{\includegraphics[width=1in,height=1.25in,clip,keepaspectratio]{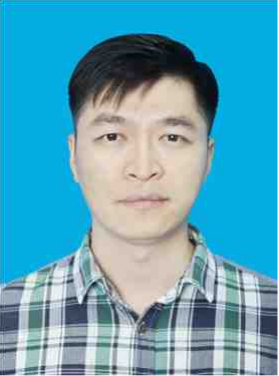}}]{Qi Liu}
is currently a Professor with the School of Future Technology at South China University of Technology. Dr. Liu received the Ph.D degree in Electrical Engineering from City University of Hong Kong, Hong Kong, China, in 2019. During 2018 - 2019, he was a Visiting Scholar at University of California Davis, CA, USA. From 2019 to 2022, he worked as a Research Fellow in the Department of Electrical and Computer Engineering, National University of Singapore, Singapore. His research interests include human-object interaction, AIGC, 3D scene reconstruction, and affective computing, etc. Dr. Liu has been an Associate Editor of the IEEE Systems Journal (2022-), and Digital Signal Processing (2022-). He was also Guest Editor for the IEEE Internet of Things Journal, IET Signal Processing, etc. He was the recipient of the Best Paper Award of IEEE ICSIDP in 2019.
\end{IEEEbiography}

\begin{IEEEbiography}[{\includegraphics[width=1in,height=1.25in,clip,keepaspectratio]{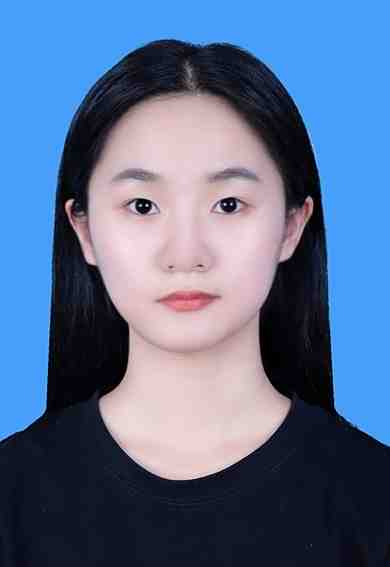}}]{Weiying Xue}
is currently pursuing an M.D degree at the School of Future Technology, South China University of Technology (SCUT), China. Her current research interests include human-object interaction, including human-object interaction detection, visual relation detection, object detection, etc.
\end{IEEEbiography}

\begin{IEEEbiography}[{\includegraphics[width=1in,height=1.25in,clip,keepaspectratio]{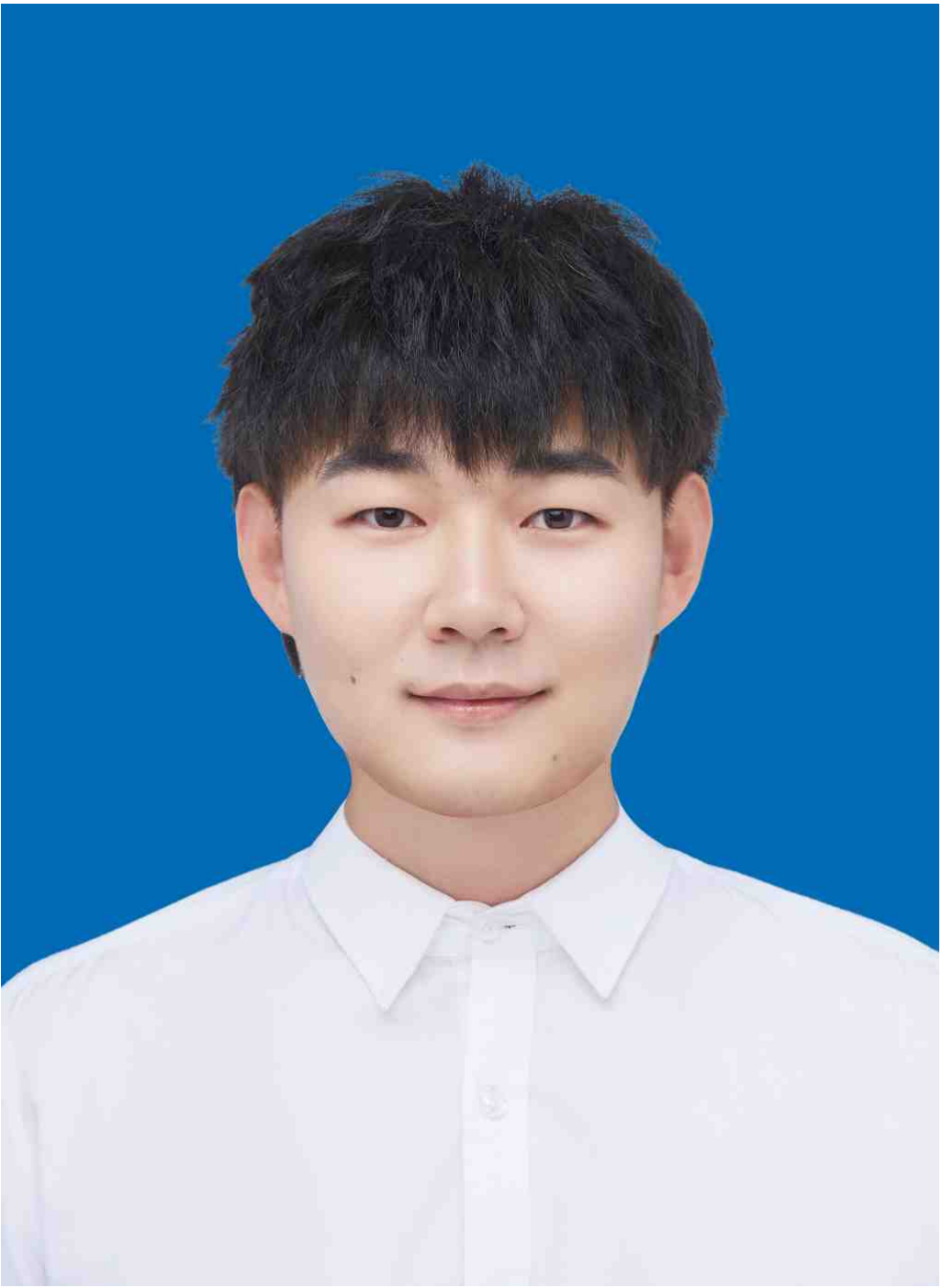}}]{Yuxiao Wang}
is currently pursuing a Ph.D degree at the School of Future Technology, South China University of Technology (SCUT), China. His research focuses on human-object interaction, including human-object interaction detection, human-object contact detection, semantic segmentation, crowd counting, etc. In addition, he has conducted research on weakly supervised learning methods.
\end{IEEEbiography}

\begin{IEEEbiography}[{\includegraphics[width=1in,height=1.25in,clip,keepaspectratio]{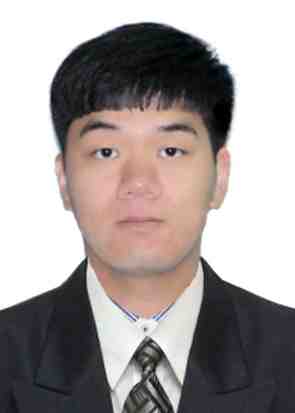}}]{Zhenao Wei}
received the D.Eng. degree from the Graduate School of Information Science and Engineering, Ritsumeikan University, Japan. He is currently a postdoctoral fellow at South China University of Technology. His research interests include game AIs, human-object interaction, and human-object contact.
\end{IEEEbiography}

\vfill

\end{document}